\pdfoutput=1

\documentclass[11pt]{article}

\usepackage[preprint]{acl}

\usepackage{times}
\usepackage{latexsym}

\usepackage[T1]{fontenc}

\usepackage[utf8]{inputenc}

\usepackage{microtype}

\usepackage{inconsolata}

\usepackage{graphicx}

\newcommand{\dataset}{\mbox{\textsc{Chatter}}}
\newcommand{\evaldataset}{\mbox{\textsc{ChatterEval}}}

\title{\textsc{{\color{blue}Ch}{\color{red}atter}}: A {\color{blue}Ch}aracter {\color{red}Attr}ibution Dataset for Narrative Understanding}

\author{Sabyasachee Baruah\textsuperscript{1} \and Shrikanth Narayanan\textsuperscript{2}\\
    Signal Analysis \& Interpretation Laboratory \\
    University of Southern California\\
    \texttt{sbaruah@usc.edu\textsuperscript{1}}, \texttt{shri@usc.edu\textsuperscript{2}}}

\begin{document}
\maketitle
\begin{abstract}
Computational narrative understanding studies the identification, description, and interaction of the elements of a narrative: characters, attributes, events, and relations.
Narrative research has given considerable attention to defining and classifying character types.
However, these character-type taxonomies do not generalize well because they are small, too simple, or specific to a domain.
We require robust and reliable benchmarks to test whether narrative models truly understand the nuances of the character's development in the story.
Our work addresses this by curating the \dataset{} dataset that labels whether a character portrays some attribute for 88124 character-attribute pairs, encompassing 2998 characters, 12967 attributes and 660 movies.
We validate a subset of \dataset{}, called \evaldataset{}, using human annotations to serve as a benchmark to evaluate the character attribution task in movie scripts.
\evaldataset{} also assesses narrative understanding and the long-context modeling capacity of language models.
\end{abstract}

\section{Introduction}

Narrativity occurs when characters interact with each other, triggering events that are temporally, spatially, and causally connected.
This sequence of events forms the story.
\citet{piper-etal-2021-narrative} provided a symbolic definition of narrativity in which they asserted that narrativity occurs when the narrator $\mathcal{A}$ tells the perceiver $\mathcal{B}$ that some agent $\mathcal{C}$ performed the action $\mathcal{D}$ on another agent $\mathcal{E}$ at place $\mathcal{F}$ and time $\mathcal{G}$ for some reason $\mathcal{H}$.
\citet{10447353} used this definition to identify four main elements of any narrative: characters, attributes, events, and relations.
\citet{labatut2019extraction} explored different types of character interactions and emphasized the central role characters play in narratives.
Characters drive the plot forward through their actions, develop attributes, arouse tension and emotion in the story by creating conflicts or bonds with other characters, and embody tropes and stereotypes to relate to the audience.
The vitality of characters in narratives makes character understanding an essential task in narrative research.

\begin{table}[t]
    \centering
    \resizebox{\columnwidth}{!}{
    \begin{tabular}{llcc}
    \hline
    Dataset & Domain & Type & Size \\
    \hline
    \citet{10.1093/llc/fqv067} & Folklore & Archetype & 282 \\
    \citet{skowron2016automatic} & Movies & Role & 2010 \\
    \citet{brahman-etal-2021-characters-tell} & Literature & Description & 9499 \\
    \citet{sang-etal-2022-mbti} & Movies & Personality & 28653 \\
    \citet{yu-etal-2023-personality} & Literature & Traits & 52002 \\
    \hline
    \dataset{} & Movies & Tropes & 88124 \\
    \evaldataset{} & Movies & Tropes & 1061 \\
    \hline
    \end{tabular}}
    \caption{Comparison with other attribution datasets in terms of the attribute type and size. Size denotes the number of character-attribute pairs. \dataset{} is the largest character attribution dataset collected so far.}
    \vspace{-17px}
    \label{tab:data}
\end{table}

Narratologists have explored various approaches to operationalize the character-understanding task.
For example, \citet{inoue-etal-2022-learning} presented character understanding as a suite of document-level tasks that included gender and role identification of the character, \textit{cloze} tasks, quote attribution, and question answering.
\citet{li-etal-2023-multi-level} adopted coreference resolution, character linking, and speaker guessing tasks, and \citet{azab-etal-2019-representing} used character relationships and relatedness to evaluate character representations.
We organized the character-understanding tasks into the following categories.
\textbf{1) Identification} tasks find the unique set of characters and their mentions.
It includes entity recognition, entity linking, and coreference resolution.
The \textbf{2) Quotation} task maps utterances to characters.
\textbf{3) Attribution} tasks, such as personality classification, persona modeling, and description generation, describe the character.
The \textbf{4) \textit{Cloze}} task asks the model to fill in the correct character name given an anonymized character description, story summary, or story excerpt.
\textbf{5) Relation} tasks, such as relation classification, draw similarities between characters.

Among these tasks, character attribution is the most challenging because there exists a multitude of ways to qualify a character, such as personality \citep{sang-etal-2022-mbti}, adjectives \citep{yu-etal-2023-personality}, persona \citep{bamman-etal-2013-learning,bamman-etal-2014-bayesian}, archetypes \citep{10.1093/llc/fqv067}, roles \citep{skowron2016automatic}, and descriptions \citep{brahman-etal-2021-characters-tell}.
Each of these approaches has drawbacks. For example, personality scales such as Big5 and MBTI cannot capture all the variation in characterization and can be difficult to interpret \citep{zillig2002we}.
Adjective descriptors are too general and apply only to a limited context in the story.
Persona roles are not explicitly defined.
Archetypes \citep{propp1968morphology,jung2014archetypes} are abstract concepts specific to the domain of interest.
Character descriptions are detailed, freeform, and scalable to any number of characters.
However, we cannot factor them into simpler components or use them to compare between different characters.

We require a robust attribution taxonomy that can scale across different narratives, characters, and domains, is well-defined and discrete for effective character comparison, and necessitates document-level understanding to model them accurately.
We developed the \dataset{} dataset to fulfill this need.
The \dataset{} dataset uses tropes from the TVTropes website as the attribute type to describe characters.
TVTropes editors and moderators comprehensively define every trope with illustrative examples from multiple media sources.
Tropes cover a wider range of character descriptions than personality types and archetypes, and require longer context-understanding, compared to traits and adjectives, to accurately ascertain if a character portrays some trope.
Unlike character descriptions, we can efficiently compare characters and their experiences using these well-defined tropes \citep{wang-etal-2021-learning-similarity}.

The \dataset{} dataset contains labels indicating whether the character portrayed the trope in the movies they appeared in.
It contains 88124 character-trope pairs.
We drew our characters primarily from Hollywood movies across various genres.
It also provides the full-length screenplays of the movies, averaging about 25K words per screenplay.
We define the character attribution task as a \textbf{binary classification task} where, given a character-trope pair and the screenplays of the movies where the character appears, the model should predict whether the character portrayed the trope.
We validate a subset of the \dataset{} data, referred to as \evaldataset{}, using human annotations to establish an evaluation benchmark for the character attribution task.
We compared the zero-shot and few-shot performance of LLMs and \dataset's labels to assess the suitability of using the \dataset{} dataset as a training set for the attribution task.
The dataset is available at \url{https://drive.google.com/drive/folders/11egMhs-zkWSASe7zJENwHg17-6VOeXDU?usp=sharing}.

\section{Data}

\subsection{Tropes}

Tropes are storytelling devices or conventions used by the writer to easily convey some story notion to the reader.
They act like narrative motifs that readers can easily recognize, saving time and effort for the writer as they can omit details the readers can infer from the trope.
For example, the \textit{AntiHero} is a very popular trope that describes a protagonist who lacks traditional heroic qualities, often cynical and flawed, yet ultimately performs heroic actions.
The character Severus Snape in the Harry Potter stories portrays the \textit{AntiHero} trope as he gives Harry a hard time but stays loyal to Dumbledore (mentor) and works secretly to defeat Voldemort (antagonist).

Tropes can also relate to inanimate objects, events, locations or the environment.
We focus only on character tropes in our work. 
The scope of a trope can vary greatly.
Some tropes like \textit{PetTheDog} (villain performing an act of kindness) are portrayed over short contexts, typically a single scene, whereas others like \textit{HiddenDepths} (revealing unexpected talents as the story progresses) are portrayed over a longer context.
The attribution model should be able to extract information from different points in the narrative and reason over it to identify the portrayed tropes.

We use the character trope labels of TVTropes\footnote{\url{https://tvtropes.org}}.
TVTropes is a community-driven website, similar to Wikipedia, that catalogs tropes with definitions and examples.
Fans of a creative work discuss and post tropes they identify in the narrative.
Each trope has a dedicated page which contains its definition, illustrations, and portrayal examples from TV shows, movies, literature, animation, video games, and print media.
TVTropes moderators ensure that the fan-edited content is correct.
We collect the trope annotations from TVTropes to build the \dataset{} dataset.

It is important to note that the character trope annotations we collect from TVTropes are those \textit{perceived} by the reader.
These might not align with the \textit{actual} tropes intended by the creator.
Since there is no quantifiable agreement on the published content, we treat the TVTropes data as a noisy source of character attribution.

\subsection{Screenplays}

We used movies as the source of our narratives.
We chose the cinematic domain over the literary domain because it had more TVTropes labels, and we supposed it would be easier to find raters more knowledgeable about movies than books. 
Additionally, movies allow us to extend the attribution task to the multimodal domain, offering more opportunities for future research directions.

We used publicly available movie screenplays from the ScriptsonScreen\footnote{\url{https://scripts-onscreen.com}} website.
Each script is mapped to an IMDB\footnote{\url{https://imdb.com}} identifier so we can uniquely identify the movie.
Most movies in our dataset are produced in the US or the UK after 1980.
The average script size is about 25K words.
We apply a named entity classifier and name alias generator to map the character names in the script to a unique character in the IMDB cast list.
We preprocess the screenplays to find scenes, dialogues and descriptions using \citeposs{baruah-narayanan-2023-character} screenplay parser.
In total, our dataset contains screenplays of 660 movies.

\subsection{\dataset}

We build the \dataset{} dataset of character-trope pairs using the tropes of TVTropes and the screenplays of ScriptsonScreen.
First, we download the movie screenplays from ScriptsonScreen, parse and map them to an IMDB page, and match the characters occurring in the document to a character in the IMDB cast list.
Second, we search for the character in TVTropes and retrieve their character page.
Third, we collect the tropes portrayed by the character from the character page, as labeled by the TVTropes community.
Fourth, we search for the trope page and fetch its definition.
We also summarize the definition using GPT-4 for the annotation task.
The average size of the trope definition and its summary is 344 and 44 words, respectively.

We need good negative samples to evaluate the specificity of the attribution model.
However, TVTropes does not provide this information because it does not label tropes \textit{not} portrayed by the character.
Instead, we analyze the trope definition to find antonym tropes and create the negative character-trope pairs.
For example, the definition of the \textit{AntiHero} trope contains the sentence --

\vspace{5px}
\texttt{...Compare and contrast this trope with its antithesis, the AntiVillain...}
\vspace{5px}

-- which indicates that the \textit{AntiVillain} trope is contrary to \textit{AntiHero}.
We search the trope definitions to retrieve opposing tropes by checking if negation words such as "contrast", "opposite" and "counterpart" exist within a five-word context window of the antonym trope mention\footnote{The full list of negation words is included in the dataset}.
For each positive character-trope pair, we create a negative pair by either -- 1) selecting an antonym trope from the trope definition or 2) choosing a trope that is absent in any of the positive character-trope pairs for that character.
It should be harder for the attribution model to distinguish antonym tropes from tropes portrayed by the character because they are much more closely related to the portrayed tropes than a trope randomly sampled with the second method.
Therefore, the former method creates hard negatives and the latter method creates soft negatives.
We randomly choose between the two methods with 50\% probability to get a good mix of hard and soft negatives.
We add the negative samples to complete the construction of the \dataset{} dataset.

\section{Evaluation Data}

We can use \dataset{} to train attribution models, but we require a more reliable dataset for evaluation.
We annotate a subset of \dataset{} using human raters and create the \evaldataset{} dataset.

\subsection{Annotation}

We sampled movies from our dataset released after 2010 and collected the corresponding character-trope pairs for annotation.
We employed workers on the Amazon MTurk\footnote{\url{https://www.mturk.com}} crowdsource annotation platform.
We selected workers with high reputation and experience\footnote{\hspace{2px}>98\% approval rate and worked on >5000 HITs} and tested them on two separate qualification tasks, each containing five questions that asked them to decide whether the character portrayed the given trope.
The task showed them the picture of the character, the movies that starred the character with Wikipedia links, a link to the fandom page\footnote{\url{https://www.fandom.com}} where the worker can read more about the character, and the summarized trope definition with a link to the trope's TVTropes page.
Appendix~\ref{sec:annotation} shows the annotation interface.
We also surveyed the workers about the movies they had watched.
We qualified 69 workers who correctly answered at least 80\% of the questions and had watched at least ten movies out of the ones sampled for evaluation.

The qualified workers labeled whether the character portrayed the trope on a Likert scale: \textit{no}, \textit{maybe no}, \textit{not sure}, \textit{maybe yes} and \textit{yes}.
To reduce the cognitive load on the workers, they annotated the character-trope pairs in batches, each batch containing characters from at most ten movies.
Before every batch, we inform the workers about the movies they will encounter to prepare them for the annotation task.
After each batch, we estimate the worker's performance by calculating the accuracy of their ratings against the \dataset{} labels and record how much time they spent on the task.
We dropped workers whose accuracy dropped below 50\% or those speeding through the samples, disqualifying them from working on future batches.
Of the 69 workers, we disqualified 10 raters and were left with 59 reliable workers.

The annotation task ran for one month between August 11 to September 19 2024.
Throughout the task, we maintained a communication channel for the workers in the TurkerNation Slack Workspace, where we responded to any questions the workers had about the task.
We pay \$1.5 for each question which turns out to be \$18/hour because the workers spent an average of 5 minutes per question.
The entire task cost us about \$10K.
We collected three annotations per character-trope pair, totaling 5683 annotations.
The Krippendorff inter-rater reliability score is 0.448, indicating moderate agreement.

\subsection{\evaldataset}

We aggregate the annotations to build the \evaldataset{} dataset.
The character-trope pairs do not all show clear agreement.
Character attribution, like sentiment analysis, is a subjective task, and whether a character portrays some trope is perceived differently by people.
We need to assign a definite binary label to each annotated sample for the character attribution task.
We also need to drop the very ambiguous samples and those with insufficient reliable ratings to ensure high quality.

The MTurk workers answer \textit{yes}, \textit{maybe yes}, \textit{not sure}, \textit{maybe no} or \textit{no} on each question.
We map this ordinal scale to a numeric range by mapping the labels to 2, 1, 0, -1 and -2, respectively.
Each character-trope pair is annotated by at most three reliable workers.
We sum the label values for each sample to get an integer score $s$ between -6 and 6.
The higher the absolute score, the greater is the agreement among the raters.
We drop samples whose absolute integer score falls below 3.
For the remaining samples, we obtain the annotation confidence $w$ by normalizing the integer score $s$ between 0 and 1: $w = (|s| - 2)/4$.
The confidence score takes values 0.25, 0.5, 0.75 and 1.
Attribution models can use these scores to weigh their evaluation metrics.
The sample gets the label $1$ (character portrays the trope) if $s > 0$, else $0$ (character does not portray the trope).
We include the annotations of the individual raters in \evaldataset{} to encourage multiannotator modeling.

\section{Data Statistics}

\begin{table}[t]
    \centering
    \resizebox{\columnwidth}{!}{
    \begin{tabular}{lcccc}
    \hline
    Dataset & Characters & Tropes & Movies & Samples \\
    \hline
    \dataset{} & 2998 & 12967 & 660 & 88124 \\
    \evaldataset{} & 271 & 896 & 78 & 1061 \\
    \hline
    \end{tabular}}
    \caption{\dataset{} and \evaldataset's sizes}
    \label{tab:statistics}
\end{table}

\begin{table}[t]
    \centering
    \resizebox{\columnwidth}{!}{
    \begin{tabular}{lcccc}
    \hline
                         & Min  & Max    & Avg      & 95\%tile \\
    \hline
        Movies/Character & 1    & 5      & 1.04     & 1.0 \\
        Tropes/Character & 1    & 1107   & 29.40    & 91.0 \\
        Words/Script     & 3051 & 87738  & 24614.98 & 34011.35 \\ 
        Tokens/Script    & 5149 & 158038 & 41952.05 & 58933.09 \\
        Words/Segment    & 3    & 28423  & 2981.28  & 9983.8 \\
        Tokens/Segment   & 4    & 43742  & 4350.89  & 14317.15 \\
    \hline
    \end{tabular}}
    \caption{\dataset{} and \evaldataset's statistics}
    \vspace{-10px}
    \label{tab:statistics2}
\end{table}

\begingroup
\renewcommand{\arraystretch}{1.4}
\begin{table*}[t]
    \small
    \centering
    \begin{tabular}{lp{4cm}p{8cm}l}
    \hline
        Character & Movies & Trope & Label \\
    \hline
        Benoit Blanc
        & \textit{Knives Out (2019)}, \textit{Glass Onion (2022)}
        & The \textbf{NiceGuy} trope describes a character who is kind, friendly, morally good, and socially pleasant.
        & 1 \\
        Mark Watney
        & \textit{The Martian (2015)}
        & The \textbf{EarnYourHappyEnding} trope involves characters enduring significant hardship, anguish, and grief, but ultimately achieving a happy ending through hard work or love
        & 1 \\
        Arthur Fleck
        & \textit{Joker (2019)}
        & The \textbf{TheDogBitesBack} trope occurs when a villain is attacked or betrayed by an abused subordinate or victim who seizes the chance for revenge
        & 0 \\
        Dr. King Schultz
        & \textit{Django Unchained (2012)} 
        & The \textbf{SoreLoser} trope describes a character who reacts to defeat with anger, accusations, and bad behavior 
        & 0 \\
    \hline
    \end{tabular}
    \caption{Character-Trope examples from the \evaldataset{} dataset}
    \vspace{-10px}
    \label{tab:examples}
\end{table*}
\endgroup

\dataset{} contains 88124 character-trope pairs with an almost 50-50 split between positive and negative pairs.
It covers 2998 unique characters from 660 movies and 12967 tropes.
\evaldataset{} contains 1061 human-annotated character-trope pairs, covering 271 characters, 896 tropes, and 78 movies.
It contains 555 positive and 506 negative pairs.
Tables \ref{tab:statistics} and \ref{tab:statistics2} show some data statistics of \dataset{} and \evaldataset{}.

Most characters appear in a single movie script in our dataset.
\dataset{} contains attribution labels for about 30 tropes for each character.
Segments indicate the portions of a movie script where a character speaks or is mentioned.
As shown in Table~\ref{tab:statistics2}, the maximum size of a movie script or a character segment can exceed 87K and 28K words, respectively.
This corresponds to 158K and 43K tokens, respectively (We used the Llama3 \citep{dubey2024llama} tiktoken-based BPE tokenizer).
Therefore, we must use a model capable of handling long contexts for the attribution task.

Most movies in our dataset have been produced in the UK or the US after 1980.
They cover a wide range of genres.
The top five genres -- \textit{Action}, \textit{Drama}, \textit{Thriller}, \textit{Adventure}, and \textit{Comedy} -- cover more than 50\% of the movies.
Table~\ref{tab:examples} shows qualitative examples from the \evaldataset{} dataset.
The tropes can relate to the character's attitude and personality (\textit{NiceGuy}), some experience or incident (\textit{SoreLoser}), or their overall story (\textit{EarnYourHappyEnding}).
The character attribution task entails that the model understands the trope definition, reads the scripts of the movies where the character appears, and, based on that, decides whether the character portrays the trope.
Appendix \ref{sec:tropes} lists the most commonly occurring tropes and their definitions to visualize the trope space.

Our dataset does not contain scripts for all the movies where the character appears.
For example, the character Arthur Fleck "Joker" has appeared in multiple movies, comics, and TV shows, but \dataset{} only contains the movie script of the \textit{Joker (2019)} movie.
The TVTropes contributors and our raters draw their knowledge of the character from all sources. 
However, the attribution model predicts the attribute labels based only on the screenplay and its pretraining data.
Therefore, intractable character-trope pairs could exist for which the model has insufficient information.
In future work, we will investigate ways to find such intractable samples in our dataset.

\section{Experiments}

\subsection{Models}

We establish baselines on \evaldataset{} using zero-shot and few-shot prompting.
We used two closed-source models, Gemini-1.5-Flash \citep{reid2024gemini} and GPT-4o-mini \citep{hurst2024gpt}, and three open-source models, Phi-3-small-7B-128k-Instruct \citep{abdin2024phi}, Llama-3.1-8B-Instruct \citep{dubey2024llama} and Mistral-Nemo-Instruct-12B-2407 \citep{jiang2023mistral}, in our experiments.
We selected these models because they can handle long contexts (128K tokens).

We experiment with four prompting strategies.
\textbf{1) Priors} -
We prompt the model with the character name, the list of movies where the character has appeared, and the trope definition.
We do not include any screenplay content and ask the model to find the attribution label based solely on its prior knowledge about the character.
\textbf{2) Script} -
We include the full movie script in the prompt.
\textbf{3) Segment (Zero-shot)} -
We include the segments of the movie script where the character speaks or is mentioned.
\textbf{4) Segment (Few-shot)} -
We include the character segments and two examples from the \evaldataset{} dataset.
We select examples of the same trope.
We also experimented with selecting random examples or examples of the same character, but they both performed worse than selecting same-trope examples.
We do not apply few-shot prompting with movie scripts because the prompt size becomes too large.
Appendix~\ref{sec:prompt} describes the prompt template we used.

For the last three settings -- \textbf{Script}, \textbf{Segment (Zero-shot)}, and \textbf{Segment (Few-shot)} -- we replaced character names in the movie scripts or the character segments with random alphanumeric identifiers.
We hypothesize that the LLM could be generating its response by recognizing the character name and using the knowledge it had accrued about the character from its pretraining datasets.
The character-anonymization step possibly prevents the LLM from relying on its prior knowledge about the character and forces it to decide the attribution label based solely on the given movie script or character segments.
We observed that prompting with the non-anonymized documents produced better performance, validating our hypothesis.
Future work should explore more effective prompting strategies to nullify the effect of the model's priors on the generated response.

We perform greedy decoding.
We also evaluate \dataset's labels against the annotations of \evaldataset{} to assess its suitability as a training set for character attribution.
We use permutation tests at $\alpha = 0.05$ to compare the performance of different prompting strategies and models.
We apply Bonferroni correction to correct for multiple comparisons.

\subsection{Results}

\begin{table}[t]
    \centering
    \resizebox{\columnwidth}{!}{
    \begin{tabular}{llcccc}
    \hline
    Prompt & Model & Acc & P & R & F1 \\
    \hline
    & Random & 50.0 & 52.3 & 50.0 & 51.1 \\
    & \dataset & \textit{80.6} & \textit{81.0} & \textit{82.2} & \textit{81.6} \\
    \hline
    Priors & Gemini-1.5 & \textbf{81.9} & \textbf{91.4} & \textbf{72.2} & \textbf{80.7} \\
    & GPT-4o & 76.2 & 98.1 & 55.6 & 71.0 \\
    & Phi-7B & 56.6 & 89.1 & 19.5 & 32.0 \\
    & Llama-8B & 71.0 & 68.9 & 81.4 & 74.6 \\
    & Mistral-12B & 73.1 & 83.0 & 61.3 & 70.5 \\
    \hline
    Script & Gemini-1.5 & 72.4 & 83.1 & 59.4 & 69.3 \\
    & GPT-4o & \textbf{73.9} & \textbf{71.2} & \textbf{84.4} & \textbf{77.2} \\
    & Phi-7B & 65.5 & 77.4 & 48.1 & 59.3 \\
    & Llama-8B & 61.4 & 61.4 & 70.7 & 65.7 \\
    & Mistral-12B & 56.1 & 60.5 & 45.7 & 52.1 \\
    \hline
    Segment & Gemini-1.5 & 73.5 & 86.6 & 58.4 & 69.8 \\
    (Zero-Shot) & GPT-4o & \textbf{78.5} & \textbf{77.4} & \textbf{83.3} & \textbf{80.3} \\
    & Phi-7B & 73.3 & 77.1 & 68.8 & 72.7 \\
    & Llama-8B & 69.8 & 65.6 & 87.7 & 75.1 \\
    & Mistral-12B & 69.3 & 81.4 & 53.2 & 64.3 \\
    \hline
    Segment & Gemini-1.5 & 65.5 & 93.4 & 36.7 & 52.7 \\
    (Few-Shot) & GPT-4o & \textbf{74.5} & \textbf{79.5} & \textbf{69.2} & \textbf{74.0} \\
    & Phi-7B & 58.4 & 58.3 & 93.2 & 71.7 \\
    & Llama-8B & 60.8 & 61.0 & 75.3 & 67.4 \\
    & Mistral-12B & 62.0 & 62.0 & 73.6 & 67.3 \\
    \hline
    \end{tabular}}
    \caption{Prior, Zero-shot, and Few-shot performance on \evaldataset{} for the character attribution binary classification task. The \textbf{\dataset{}} row uses the \dataset's labels as predictions. We bolden the model row with the best accuracy (or F1) for each prompting strategy.}
    \vspace{-15px}
    \label{tab:baseline}
\end{table}

Table~\ref{tab:baseline} shows the performance of the different prompting strategies.
The closed-sourced models were significantly more accurate than the open-sourced models for all strategies.
This is expected because the closed-sourced models supposedly contain more parameters and have been trained on more extensive data.
Gemini-1.5 has the strongest prior knowledge, followed by GPT-4o.
They performed equally well when we prompted them on the full movie script.
However, GPT-4o's zero-shot and few-shot accuracy on character segments was significantly better than Gemini's.
There was no significant difference between the accuracy scores of Phi-7B, Llama-8B, and Mistral-12B for any of the prompting strategies, except for priors where Phi-7B performed worse than random.

Comparing the different prompting strategies, we observed that the zero-shot accuracy on character segments was significantly better than the few-shot accuracy across all the models.
This discrepancy in performance is surprising because prompting models usually provide better results with exemplars in the prompt.
A possible reason could be that the two examples in the prompt are insufficient to represent the task adequately.
Increasing the number of prompts is not viable because of the context limit.
A possible solution could be summarizing the character segments to fit more examples in the prompt.

Prompting the character segments usually performed as well or better than prompting the full movie scripts.
However, in most cases, the model's prior performance already showed strong results.
This confirms that these models were probably pretrained on the tropes data from TVTropes, making it harder to evaluate the true character attribution capability of the model.
The huge gap in performance between the priors of the closed-source models and the zero-shot or few-shot prompting results of the open-sourced models shows that there is much scope for improvement for the open-sourced models.
This is important because, given a movie script (or any other narrative document), we do not always want to prompt it using commercial APIs because they might contain private or unpublished information.
Training local attribution models on the character-trope pairs of \dataset{} should narrow this performance gap.

There is no significant difference between the results of the strongest-performing model and using the labels of the \dataset{} dataset as predictions.
Therefore, \dataset{} can serve as a good training source for the character attribution task.
The strong zero-shot performance of the closed-sourced models suggests that we can use them to create good-quality synthetic training data.

We also observe that the recall scores of the models are usually lower than their precision.
A possible reason could be the sizeable prompt size, which makes it difficult for the model to pinpoint the relevant sections from the script or the character segments.
While training character attribution models, we should further preprocess the data for more effective learning.

\section{Related Work}

Several past studies have curated character attribution datasets for narrative understanding.
\citet{10.1093/llc/fqv067} annotated seven character archetypes in Russian folktales to create the ProppLearner corpus.
\citet{skowron2016automatic} labeled hero, antagonist, sidekick, spouse and supporting roles in action-genre movies.
\citet{brahman-etal-2021-characters-tell} collected character descriptions from online study guides such as LitChart and Sparknotes, and created the LiSCU dataset for the character identification and description generation task.
\citet{sang-etal-2022-mbti} curated MBTI personality types from the personality database for movie characters.
\citet{yu-etal-2023-personality} annotated reader's notes for character traits in Chinese-translated Gutenberg books.
\citet{10447353} annotated screenplay excerpts for character attributes and evaluated in-context and chain-of-thought learning methods on the attribution task.
Table~\ref{tab:data} compares these datasets against \dataset{} and \evaldataset{}.

\section{Conclusion}

We proposed the \dataset{} and \evaldataset{} datasets for the narrative character attribution task.
We addressed the limitations of previous datasets by curating a resource that is scalable, generalizable, well-defined and discrete.
Experiments showed that \dataset{} can serve as a reliable training set and \evaldataset{} can be used as the evaluation benchmark for character attribution modeling.
Future work includes developing character attribution models on our datasets to aid creators and writers in analyzing their narratives.

\section{Limitations}

We define the character attribution task as a binary classification task, where, given the character-trope pair and the screenplays of the movies in which the character appeared, the model should predict whether or not the character portrayed the trope.
This formulation has some limitations.
First, the screenplay's narrative may not exactly resemble the story told by the movie because of tweaks made during the filming process.
Publicly available screenplays are rarely the final script but drafts from earlier in the production stage.
Second, movies could have visual cues like the non-verbal behavior of the character that are missed by the screenplay text.
Lastly, a character can appear in multiple movies, and our dataset does not contain scripts for all of them.
These limitations have important implications because the TVTropes contributors and our raters draw their knowledge of the character from all sources. 
In contrast, the attribution model predicts the attribute labels based only on the screenplay.
There could be character-attribute pairs for which the model has insufficient information and could lead to poor recall, as shown by the performance of the models in Table~\ref{tab:baseline}.

\bibliography{main}

\newpage

\appendix

\begin{figure*}[th]
    \centering
    \frame{\includegraphics[width=\textwidth]{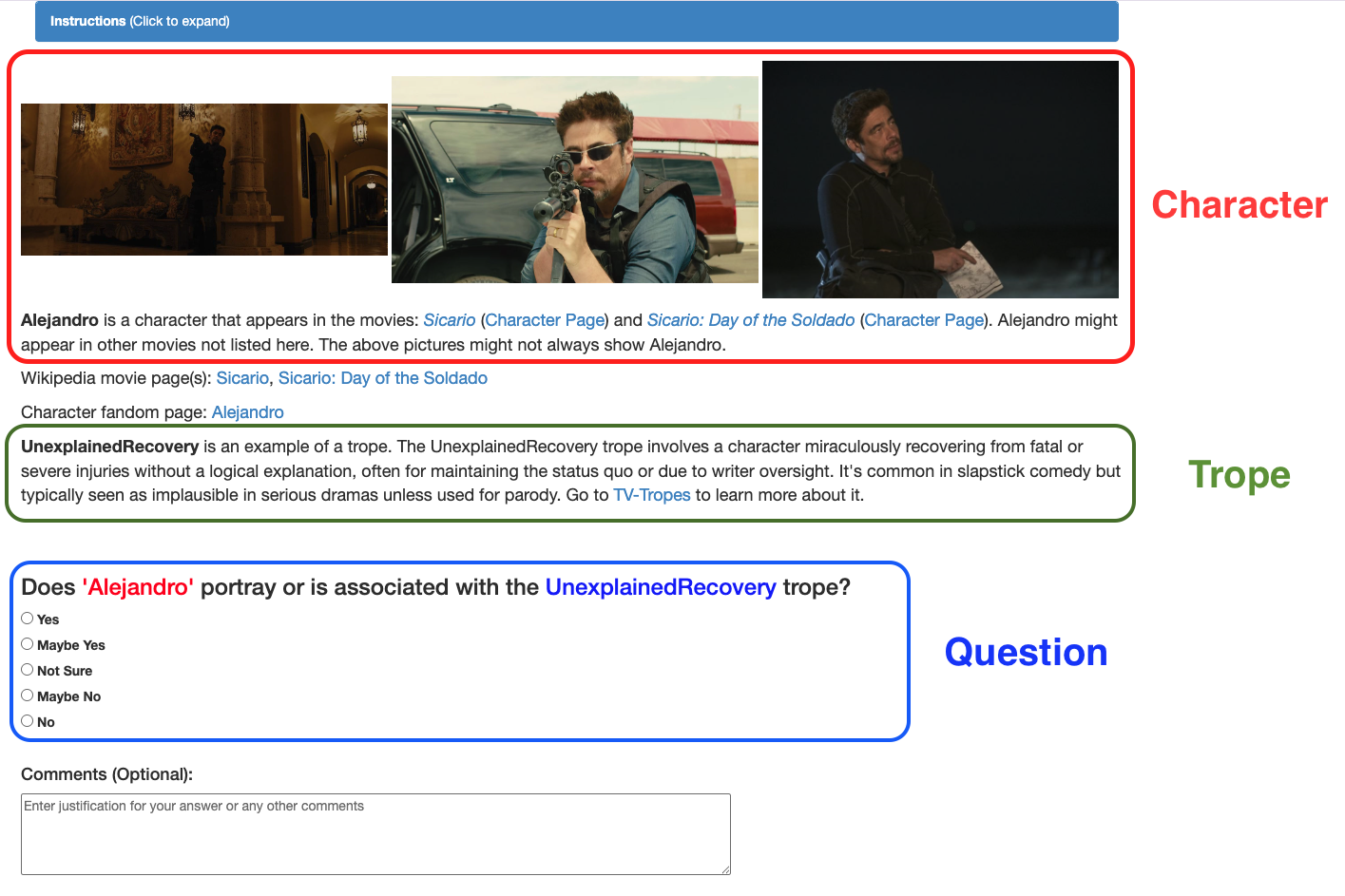}}
    \caption{Interface of the annotation task.}
    \label{fig:annotation}
\end{figure*}

\section{Annotation}
\label{sec:annotation}

We give the following annotation instructions to the Amazon MTurk workers.

\begin{verbatim}
A trope is a storytelling device or 
convention the storyteller uses to 
describe situations the audience can 
easily recognize, often used to 
stereotype characters. Think of them 
as character attributes.

Your task is as follows:

1. Identify the movie character from the 
   pictures and the given movie(s). If 
   you are having trouble recognizing 
   the character, click on the Character 
   Page link(s).

2. Read the definition of the character 
   trope. You can open the link to know 
   more.

3. Choose yes, maybe yes, not sure, 
   maybe no, or no to answer whether the 
   character portrays or is associated with 
   the trope.

4. (Optional) Explain your choice in the 
   provided text area or leave any other 
   comment behind.
\end{verbatim}

Figure~\ref{fig:annotation} shows the annotation interface that the workers use to label character attribution.

\section{TVTropes}
\label{sec:tvtropes}

TVTropes is licensed under a Creative Commons Attribution-NonCommercial-ShareAlike 3.0 Unported License.
This license allows for the copy and redistribution of the material in any medium or format for non-commercial purposes.
\dataset{} is published for academic research purposes and does not infringe the TVTropes license.

\section{ScriptsonScreen}
\label{sec:scriptsonscreen}

ScriptsonScreen aggregates movie scripts from different websites such as IMSDB\footnote{\url{https:https://imsdb.com/}}, Dailyscripts\footnote{\url{https://www.dailyscript.com/}} and Script Slug\footnote{\url{https://www.scriptslug.com/}}.
All the movie scripts we use are licensed for fair use and available for public download.

\section{TROPES}
\label{sec:tropes}

There are 445 tropes in total that have at least one character portraying them in our dataset.
We could not figure out a systematic way to cluster them because the tropes are very different from each other.
Therefore, we simply list the most common tropes and their abbreviated definition.
Tables \ref{tab:tropes1} and \ref{tab:tropes2} list the tropes that are portrayed by at least two characters in our dataset.

\begin{table*}[ht]
    \centering
    \resizebox{\textwidth}{!}{
    \begin{tabular}{lll}
        \hline
        & Trope & Definition \\
        \hline
        1. & AdaptationalAttractiveness & Plain characters are portrayed as attractive in adaptations \\
        2. & AdaptationPersonalityChange & Character personality changes during medium adaptations \\
        3. & AffablyEvil & Charming villain with kindness despite malevolent intentions \\
        4. & AntiVillain & Heroic goals achieved through questionable or evil means \\
        5. & AudienceSurrogate & Character audience identifies with for sympathy and relatability \\
        6. & BadLiar & Character fails to lie convincingly; creates transparent falsehoods \\
        7. & BerserkButton & Character's minor trigger causes explosive anger reaction \\
        8. & BitchInSheepsClothing & Deceptive character appears kind but is secretly villainous \\
        9. & BlueIsHeroic & Blue signifies heroism through calm, disciplined characters \\
        10. & CatchPhrase & Repetitive distinctive phrase by a character or category \\
        11. & ChildrenAreInnocent & Children embody innocence and purity, contrasting adult corruption \\
        12. & Cloudcuckoolander & Cheerfully eccentric character, detached from reality, unexpectedly wise \\
        13. & ControlFreak & Obsessively enforces rules, hindering effectiveness and dissent \\
        14. & CynicismCatalyst & Trauma shifts idealistic character to cynicism and moral ambiguity \\
        15. & DarkAndTroubledPast & Character's tragic past shapes their personality and behavior \\
        16. & DeadpanSnarker & Sarcastic character who critiques and deflates others' egos \\
        17. & Determinator & Unyielding persistence towards goals, regardless of challenges \\
        18. & DramaQueen & Excessively dramatic characters who overreact and seek attention \\
        19. & EvenEvilHasStandards & Villain shows moral boundaries despite overall remorselessness \\
        20. & EveryoneHasStandards & Characters uphold personal standards despite their own flaws \\
        21. & EvilWearsBlack & Villains wear black, symbolizing darkness and aggression \\
        22. & FauxAffablyEvil & Polite villains hiding true cruelty for manipulation and enjoyment \\
        23. & Gaslighting & Manipulating perception to induce doubt and confusion \\
        24. & GentleGiant & Big, kind character defying intimidating appearance; gentle and reliable \\
        25. & GoingNative & Character embraces new culture, rejects original society \\
        26. & GrinOfRage & Smiling while angry, used for intimidation or provocation \\
        27. & GuileHero & Cunning hero uses wit and charm for noble goals \\
        28. & HairTriggerTemper & Explosive anger at minor irritations; unpredictable and dangerous \\
        29. & HeroAntagonist & Good antagonist opposing protagonist for noble reasons \\
        30. & HiddenDepths & Characters unveil unexpected talents, deepening their complexity \\
        31. & Hypocrite & Authority figures failing to uphold their own ideals \\
        32. & ItsAllAboutMe & Self-centered character believes world revolves around them \\
        33. & Jerkass & Self-centered character creating conflict for comedic/dramatic effect \\
        34. & JerkassHasAPoint & Morally flawed character speaks uncomfortable but true points \\
        35. & KarmaHoudiniWarranty & Villain faces late justice, satisfying audience's desire for retribution \\
        36. & KickTheDog & Character's cruel act establishes evil, shifts audience sympathy \\
        37. & LackOfEmpathy & Characters recognize emotions but lack emotional connection \\
        38. & LargeHam & Flamboyant, over-the-top character adding drama and charisma \\
        39. & LaserGuidedKarma & Immediate consequences for characters' actions reinforce moral lessons \\
        40. & LivingMacGuffin & Person drives quests due to intrinsic value or attributes \\
        41. & LoveAtFirstSight & Instant deep love between characters upon first meeting \\
        42. & LoveInterest & Romantic character involved with another, often archetypal roles \\
        43. & ManlyTears & Stoic male character cries from strong, dignified emotions \\
        44. & MeaningfulName & Character names reflect traits or roles meaningfully \\
        45. & MoralityPet & Villain's bond with innocent character prompts redemption \\
        46. & MorphicResonance & Characters retain recognizable traits across different forms \\
        47. & MyGodWhatHaveIDone & Character regrets harmful actions, prompting remorse and conflict \\
        48. & Narcissist & Character obsessed with self-admiration and validation, often hostile \\
        49. & NeverBareheaded & Character always wears headgear, never seen bare-headed \\
        50. & NiceGirl & Kind, friendly character contrasting cynical figures; endearing presence \\
        \hline
    \end{tabular}}
    \caption{Tropes and their definitions (first part)}
    \label{tab:tropes1}
\end{table*}

\begin{table*}[ht]
    \centering
    \resizebox{\textwidth}{!}{
    \begin{tabular}{lll}
        \hline
        & Trope & Definition \\
        \hline
        51. & NiceGuy & Kind, morally good character contrasting with cynicism \\
        52. & NiceJobBreakingItHero & Hero's victory causes unintended negative consequences \\
        53. & NoCelebritiesWereHarmed & Parody characters resembling real celebrities, often with altered names \\
        54. & NoSocialSkills & Characters lacking social awareness, often blunt but intelligent \\
        55. & OhCrap & Character realizes impending disaster, leading to panic or horror \\
        56. & OnlyOneName & Characters known by only one name, no identifiers \\
        57. & OnlySaneWoman & Rational character amidst chaotic, irrational peers; often frustrated \\
        58. & PayEvilUntoEvil & Revenge-driven morality blurs hero-villain boundaries \\
        59. & PosthumousCharacter & Dead character influences plot through memories or narratives \\
        60. & ReasonableAuthorityFigure & Open-minded leader who evaluates heroes' claims rationally \\
        61. & RebelliousSpirit & Character defies norms, follows personal rules, often anti-heroic \\
        62. & ShadowArchetype & Character reflecting protagonist's denied flaws, causing conflict and growth \\
        63. & SirSwearsALot & Character known for excessive swearing, revealing deeper traits \\
        64. & StepfordSmiler & Cheerful facade hides inner turmoil and psychological issues \\
        65. & TheChessmaster & Cunning strategist who manipulates events for personal gain \\
        66. & TheDon & Ruthless crime patriarch with moral codes, protective yet shrewd \\
        67. & TheHeart & Caretaker and moral compass of the team \\
        68. & TheImmune & Character immune to disease, pivotal for finding cure \\
        69. & TheJerkIndex & Characters exhibiting rudeness contrasting with polite behavior \\
        70. & TheSociopath & Character lacking empathy, manipulative, and morally ambiguous \\
        71. & TopHeavyGuy & Exaggerated large upper body, skinny legs \\
        72. & TragicVillain & Sympathetic villain regrets their actions, seeks redemption \\
        73. & UncertainDoom & Ambiguous fate of a character, survival uncertain \\
        74. & UndignifiedDeath & Ridiculous, embarrassing deaths blending humor and tragedy \\
        75. & UnwittingInstigatorOfDoom & Unintentional catalyst for disaster, unaware of their role \\
        \hline
    \end{tabular}}
    \caption{Tropes and their definitions (second part)}
    \label{tab:tropes2}
\end{table*}

\section{Prompt}
\label{sec:prompt}

We used the following prompt template in our experiments.

\begin{verbatim}
A character trope is a story-telling 
device used by the writer to describe 
characters.

Given below is the definition of the 
$TROPE$ trope enclosed between the 
tags <TROPE> and </TROPE>.
Following that is a movie script 
enclosed between the 
tags <SCRIPT> and </SCRIPT>. 
The character "$CHARACTER$" appears 
in the movie script.

Read the movie script carefully 
and based on that answer yes or no
if the character "$CHARACTER$" 
portrays or is associated with the 
$TROPE$ trope.
If yes, give a brief explanation.
Answer based only on the movie script.
Do not rely on your prior knowledge.

<TROPE>
$DEFINITION$
</TROPE>

<SCRIPT>
$SCRIPT$
</SCRIPT>

Does the character "$CHARACTER$" 
portray or is associated with 
the $TROPE$ trope in the above 
movie script?
Answer yes or no. 
If yes, give a brief explanation. 
Do not use MarkDown.
\end{verbatim}

We replace \$CHARACTER\$, \$TROPE\$, \$DEFINITION\$ and \$SCRIPT\$ with the character name, trope name, definition of the trope and the movie script during prompting.

\end{document}